\documentclass[11pt,a4paper]{article}
\usepackage[hyperref]{naaclhlt2019}
\usepackage{times}
\usepackage{latexsym}
\usepackage{url}
\usepackage{graphicx}
\usepackage{amsmath}
\usepackage{enumerate}
\usepackage{amsfonts}
\usepackage{array}
\usepackage{subfigure}
\usepackage{amssymb}
\usepackage{multirow}
\usepackage{booktabs}
\usepackage{footmisc}
\usepackage{floatrow}
\usepackage{mdwlist}
\usepackage{xcolor}
\usepackage{paralist}
\usepackage[linesnumbered,ruled,vlined]{algorithm2e}

\aclfinalcopy 
\def\aclpaperid{1655} 

\definecolor{darkgreen}{RGB}{84,174,50}
\newcommand{\up}[1]{$\textcolor{darkgreen}{(#1)}$}
\newcommand{\dn}[1]{$\textcolor{red}{(#1)}$}
\newcommand{\en}{$\textcolor{white}{(+0.00)}$}

\title{Extract and Edit: An Alternative to Back-Translation for \\
Unsupervised Neural Machine Translation
}

\author{Jiawei Wu \and Xin Wang \and William Yang Wang \\
Department of Computer Science\\
University of California, Santa Barbara\\
Santa Barbara, CA 93106 USA\\
{\tt \{jiawei\_wu,xwang,william\}@cs.ucsb.edu}}

\date{}

\begin{document}
\maketitle
\begin{abstract}
The overreliance on large parallel corpora significantly limits the applicability of machine translation systems to the majority of language pairs. Back-translation has been dominantly used in previous approaches for unsupervised neural machine translation, where pseudo sentence pairs are generated to train the models with a reconstruction loss. However, the pseudo sentences are usually of low quality as translation errors accumulate during training. To avoid this fundamental issue, we propose an alternative but more effective approach, \textbf{extract-edit}, to extract and then edit real sentences from the target monolingual corpora. Furthermore, we introduce a comparative translation loss to evaluate the translated target sentences and thus train the unsupervised translation systems. Experiments show that the proposed approach consistently outperforms the previous state-of-the-art unsupervised machine translation systems across two benchmarks (English-French and English-German) and two low-resource language pairs (English-Romanian and English-Russian) by more than $2$ (up to $3.63$) BLEU points.
\end{abstract}

\section{Introduction}
\label{sec:intro}
Promising results have been achieved in Neural Machine Translation (NMT) by representation learning~\cite{cho2014learning,sutskever2014sequence}.
But recent studies~\cite{koehn2017six,isabelle2017challenge,sennrich2016grammatical} highlight the overreliance of current NMT systems on large parallel corpora.
In real-world cases, the majority of language pairs have very little parallel data, so the models need to leverage monolingual data to address this challenge~\cite{gulcehre2015using,zhang2016exploiting,he2016dual,yang2018unsupervised}.

\begin{figure}[t]
\centering
\includegraphics[width=1\textwidth]{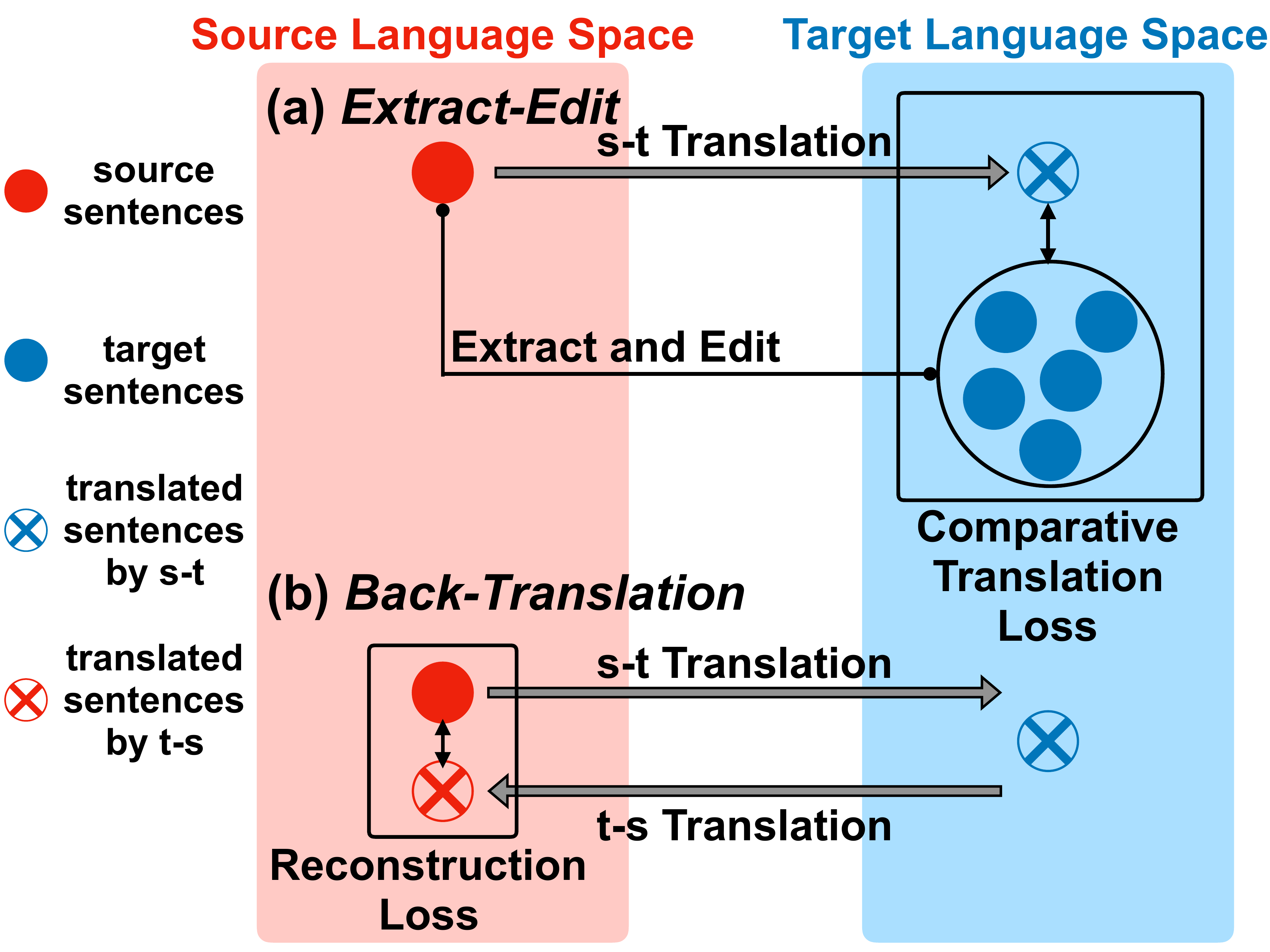}
\caption{The comparison between two approaches of unsupervised NMT, extract-edit and back-translation. When training the source-to-target (s-t) translation model, instead of using the t-s back-translated sentences to train the model, we directly set the extracted-edited sentences as pivotal points to guide the training.}
\label{fig:compare}
\end{figure}

While many studies have explored how to use the monolingual data to improve translation performance with limited supervision, latest approaches~\cite{lample2017unsupervised,artetxe2017unsupervised,lample2018phrase} focus on the fully unsupervised scenario. Back-translation has been dominantly used in these approaches, where pseudo sentence pairs are generated to train the translation systems with a reconstruction loss. However, it is inefficient because the generated pseudo sentence pairs are usually of low quality.
During the dual learning of back-translation, the errors could easily accumulate and thus the learned target language distribution would gradually deviate from the real target distribution. This critical drawback hinders the further development of the unsupervised NMT systems.

An alternative solution is to extract real parallel sentences from comparable monolingual corpora, and then use them to train the NMT systems. Recently, neural-based methods~\cite{chu2016parallel,grover2017bilingual,gregoire2018extracting} aim to select potential parallel sentences from monolingual corpora in the same domain. However, these neural models need to be trained on a large parallel dataset first, which is not applicable to language pairs with limited supervision.

In this paper, we propose a radically different approach for unsupervised NMT---\textbf{extract-edit}, a powerful alternative to back-translation (see \autoref{fig:compare}).
Specifically, to train the source-to-target translation model, 
we first extract potential parallel sentence candidates in the target language space given a source language sentence.
Since it cannot be guaranteed that there always exist potential parallel sentence pairs in monolingual corpora, we further propose a simple but effective editing mechanism to revise the extracted sentences, making them aligned with the source language sentence.
Then a comparative translation loss is introduced to evaluate the translated sentence based on the extracted-and-edited ones and train the translation model.
Compared to back-translation, extract-edit avoids the distribution deviation issue by extracting and editing real sentences from the target language space. 
Those extracted-and-edited sentences serve as pivotal points in the target language space to guide the unsupervised learning. Thus, the learned target language distribution could be closer to the real one.
The extract-edit model and the translation model, the two major parts of our method, can be jointly trained in a fully unsupervised way.

Empirical results on popular benchmarks show that exact-edit consistently outperforms the state-of-the-art unsupervised NMT system~\cite{lample2018phrase} with back-translation across four different languages pairs. In summary, our main contributions are three-fold\footnote{The source code can be found in this repository: \url{https://github.com/jiaweiw/Extract-Edit-Unsupervised-NMT}}:
\begin{itemize}
\item We propose a more effective alternative paradigm to back-translation, extract-edit, to train the unsupervised NMT systems with potentially real sentence pairs;
\item We introduce a comparative translation loss for unsupervised learning, which optimizes the translated sentence by maximizing its relative similarity with the source sentence among the extracted-and-edited pairs;   
\item Our method advances the previous state-of-the-art NMT systems across four different language pairs under monolingual corpora only scenario.
\end{itemize}

\section{Background}
\label{sec:background}
Without parallel sentence pairs as constraints on mapping language spaces, training NMT systems is an ill-posed problem because there are many potential mapping solutions. 
Nevertheless, some promising methods have been proposed in this field~\cite{lample2017unsupervised,artetxe2017unsupervised,lample2018phrase}. 
The main technical protocol of these approaches can be summarized as three steps: \textit{Initialization}, \textit{Language Modeling}, and \textit{Back-Translation}. 
In this section, we mainly introduce the three steps and the crucial settings that we have followed in our work. 

In the remainder of the paper, we denote the space of source and target languages by $\mathcal{S}$ and $\mathcal{T}$, respectively. $enc$ and $dec$ refer to the encoder and decoder models in the sequence-to-sequence systems. $V_{s\to t}$ stands for the composition of $enc$ in the source language and $dec$ in the target language, which can be viewed as the source-to-target translation system.

\paragraph{Initialization}
\label{subsec:init}
Given the ill-posed nature of the unsupervised NMT task, a suitable initialization method can help model the natural priors over the mapping of two language spaces we expect to reach. There are mainly two initialization methods: (1) bilingual dictionary inference~\cite{conneau2017word,artetxe2017unsupervised,lample2017unsupervised} and (2) byte-pair encoding (BPE)~\cite{sennrich2015neural,lample2018phrase}. As shown in~\newcite{lample2018phrase}, the inferred bilingual dictionary can provide a rough word-by-word alignment of semantics, and the BPE can reduce the vocabulary size and eliminate the presence of unknown words in the output results.

In our \emph{extract-edit} approach, to extract potential parallel sentence pairs, we need to compare the semantic similarity of sentences between two languages first. A proper initialization can also help align the semantic spaces and extract potential parallel pairs within them. Thus, following the previous methods, we use the inferred bilingual dictionary as described in~\newcite{conneau2017word} for unrelated language pairs and the shared BPE in~\newcite{lample2018phrase} as initialization for related ones.

\begin{figure*}[t]
\centering
\includegraphics[width=1\textwidth]{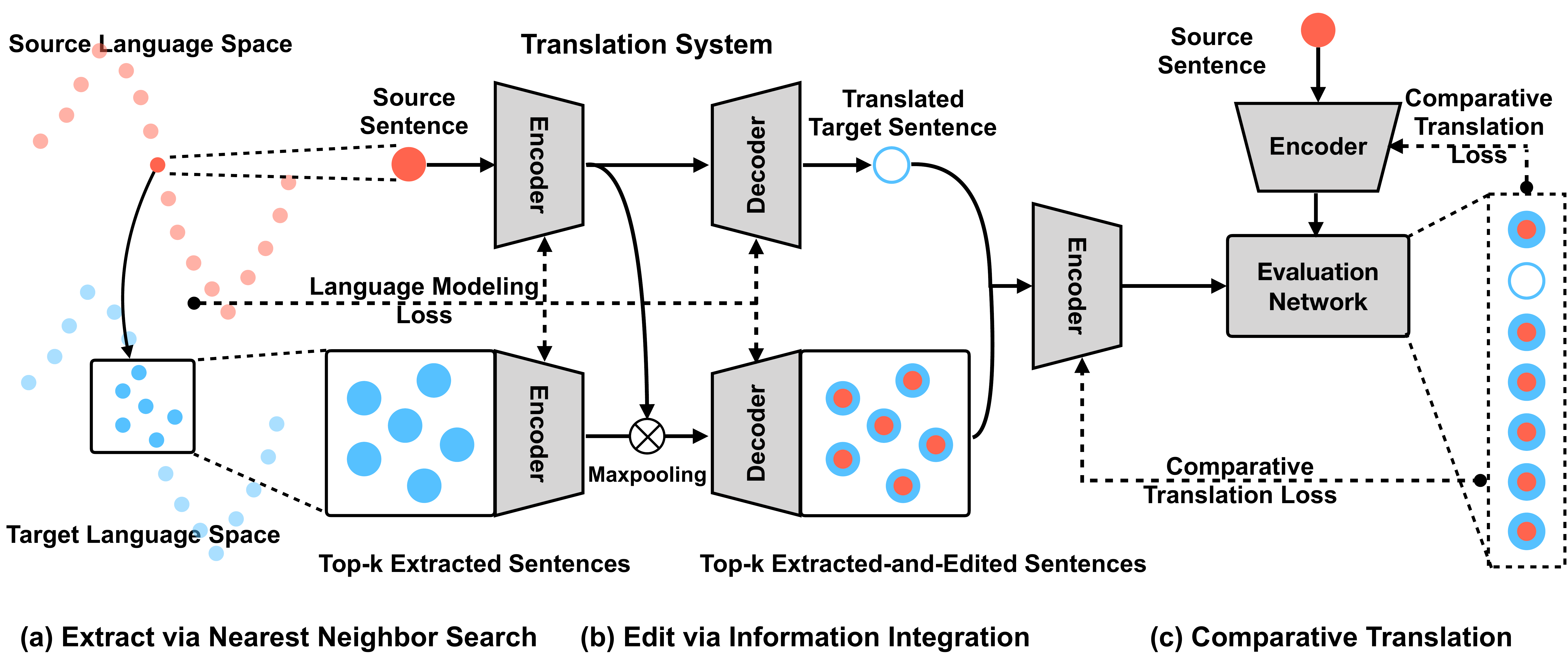}
\caption{The overview of our unsupervised NMT model based on the \emph{extract-edit} approach.
Given a source sentence, (a) the top-$k$ potential parallel sentences of the target language are extracted via nearest neighbor search. (b) The extracted sentences are further edited with the source sentence. 
(c) The evaluation network evaluates the translated sentence and the extracted-and-edited sentences based on their similarities with the source sentence.
Note that (1) all the encoders share the same parameters (same for decoders); (2) the decoding processes are non-differentiable, so the language modeling loss and the comparative translation loss are used to train the learning modules before and after the decoding processes, respectively.}
\label{fig:overview}
\end{figure*}

\paragraph{Language Modeling}
\label{subsec:lm}
After a proper initialization, given large amounts of monolingual data, we can train language models on both source and target languages. These models express a data-driven prior about the composition of sentences in each language. In NMT, language modeling is accomplished via denosing autoencoding, by minimizing:
\begin{equation}
\begin{split}
\label{equ:lm}
\mathcal{L}_{lm}(\theta_{enc}, \theta_{dec}) = & \mathbb{E}_{x\sim \mathcal{S}}[-\log V_{s\to s}(x|C(x))] + \\
         & \mathbb{E}_{y\sim \mathcal{T}}[-\log V_{t\to t}(y|C(y))]
\end{split}
\end{equation}
where $C$ is a noise model with some words dropped and swapped, $\theta_{enc}$ and $\theta_{dec}$ are the learnable parameters of $enc$ and $dec$. 
$V_{s\to s}$ and $V_{t\to t}$ are the encoder-decoder language models on the source and target sides, respectively.

In our \emph{extract-edit} approach, we follow similar settings and adopt the noise model proposed by~\newcite{lample2017unsupervised}. Note that the parameters of all $enc$ are shared (same for $dec$) in our framework to ensure a strong alignment and mapping between two languages. 
This sharing operation is essential for both the translation model and the extract-edit model. 
Thus, we use $enc$ to represent encoders in source language modeling $enc_{\mathcal{S}}$ and in target language modeling $enc_{\mathcal{T}}$ (same for $dec$).

\paragraph{Back-Translation}
\label{subsec:backtrans}
Back-translation~\cite{sennrich2015neural} has been dominantly used in prior work to train the unsupervised NMT system. It couples the source-to-target translation model with a backward target-to-source model and trains the whole system with a reconstruction loss. This can be viewed as converting the unsupervised problem into a supervised scenario by generating pseudo language pairs~\cite{he2016dual}.

Despite the popularity of back-translation in the previous methods~\cite{lample2017unsupervised,artetxe2017unsupervised,lample2018phrase}, we argue that it suffers from the low-quality pseudo language pairs. Thus, in this work, we propose a new paradigm, extract-edit, to address this issue by extracting and editing potential real parallel sentences. 
Below we describe our approach in details.

\section{Extract-Edit}
\label{sec:method}
The overview of our \emph{extract-edit} approach is shown in Figure~\ref{fig:overview}. We first extract and edit real sentences from the target language space according to their similarities with the source sentence. These extracted-and-edited sentences serve as pivotal points in the target language space, which locate a probable region where the real target sentence could be.
Then we introduce a comparative translation loss to evaluate the translated sentence and train the system. 
Basically, the comparative translation loss encourages the translated sentence to approximate the real sentence by maximizing its relative similarity with the source sentence compared to the extracted-and-edited sentences. 
As a result, we manage to minimize the deviation of the learned target language distribution and the mapping noises between two language spaces.

\subsection{Extract}
\label{subsubsec:extract}
Most existing methods in comparable corpora mining introduce two encoders to represent sentences of two languages separately, and then use another network to measure the similarity~\cite{chu2016parallel,grover2017bilingual,gregoire2018extracting}. 
However, owing to the shared encoders and decoders in language modeling, the semantic spaces of two languages are already strongly connected in our scenario.

Therefore, to avoid extra computation resources, we directly use the $enc$ in language modeling to obtain sentence embeddings for two languages. 
As shown in Figure~\ref{fig:overview}~(a), for a given source sentence $s$, we use the nearest neighbor search based on $L_2$ distance to find top-$k$ real sentences from the target language space ($k$ is a hyper-parameter decided empirically). The sentence embeddings used for searching are computed based on the shared encoder $enc$.
The reason to choose top-$k$ sentences rather than top-$1$ is to keep a high recall rate and obtain more related samples from the target language space. Finally, given the source sentence $s$, we denote $M$ as a set of the $k$ potential parallel target sentences:
\begin{equation}
M = \{t|\min_{1,\cdots,k}(||e_s-e_t||),t\in\mathcal{T}\},
\end{equation}
where $e_s$ and $e_t$ are sentence embeddings encoded by the shared encoder $enc$.

\subsection{Edit}
\label{subsubsec:edit}
Even though the extracted sentences could serve as pivotal points to guide NMT, there is no guarantee that there always exists a parallel sentence in the target corpus.
Thus, in order to make it closer to the real paired sentence in the target language space, we propose an editing mechanism to revise the extracted target sentence $t\in M$ based on the semantics of the source sentence $s$.
As described in Figure~\ref{fig:overview} (b), we employ a maxpooling layer to reserve the more significant features between the source sentence embedding $e_s$ and the extracted sentence embedding $e_t$ ($t \in M$), and then decode it into a new sentence $t'$:
\begin{equation}
M' = \{t'|t' = dec(\text{maxpooling}(e_s,e_t)), t\in M\},
\end{equation}
where $M'$ is the set of the extracted-and-edited sentences. Based on the semantic information of the source sentence $s$, we can further improve the extracted results with this editing mechanism. Unlike other studies using the editing to generate more structural sentences~\cite{guu2018generating,hashimotoretrieve}, here the revised sentences are designed to serve as better pivotal points in the target language space to guide the translation procedure. This can also be viewed as adding constraints when aligning the two language spaces.

\subsection{Evaluate}
\label{subsubsec:eval}
Given a source sentence $s$, we can translate it as $t^*$ using the source-to-target translation model $P_{s\to t}$. Meanwhile, a set $M'$ of $k$ sentences can also be generated by the extract-edit approach described above.
Although the $M'$ may contain potential parallel sentences $t'$ for $s$, we cannot directly use $(s,t')$ as ground-truth sentence pairs to train the translation model $V_{s\to t}$ because the NMT system is sensitive to noises~\cite{cho2014properties,cheng2018towards}. The rough operation like this will result in sub-optimal translation performance.

Therefore, in order to assess the quality of the translated sentence $t^*$ and train the translation model $V_{s\to t}$, we introduce an evaluation network $R$ for evaluating the relative similarities between the source and target sentences among all sentence pairs. 
The evaluation network $R$ is a multilayer perceptron; it takes the target sentence embedding $e_t$ and source sentence embedding $e_s$ as inputs, and converts them into the joint embedding space as $r_t$ and $r_s$. So the similarity
\begin{equation}
\label{equ:cosine}
\alpha(t|s) = cosine(r_t, r_s) = \frac{r_t\cdot r_s}{||r_t||||r_s||}.
\end{equation}
Then, a softmax-like formulation is used to compute the ranking score for the translated sentence $t^*$ given the extracted-and-edited sentence set $M'$:
\begin{equation}
\label{eq:rank}
P(t^*|s,M') = \frac{exp(\lambda\alpha(t^*|s))}{\sum_{t'\in M'\cup\{t^*\}}exp(\lambda\alpha(t'|s))},
\end{equation}
where the hyper-parameter $\lambda$ is similar to the inverse temperature of the softmax function. Lower $\lambda$ encourages the model to treat all extracted-edited sentences equally, while higher $\lambda$ highlights the importance of sentences with higher-score.

\subsection{Learning}
\paragraph{Comparative Translation}
As introduced above, the ranking score calculates the relative similarity between the $<s, t^*>$ pair and all the extracted-and-edited pairs $<s, t'>$.
Assuming we have a good evaluation network $R$ with $\theta_R$ denoting its parameters, we further introduce the comparative translation loss $\mathcal{L}_{com}$ for unsupervised machine translation:
\begin{equation}
\mathcal{L}_{com}(\theta_{enc}|\theta_{R}) = -\mathbb{E}(\log P(t^*=V_{s\to t}(s)|s,M')),
\end{equation}
where $\theta_{enc}$ is the parameters of the shared encoder $enc$.
Basically, the translation model is trying to minimize the relative distance of the translated sentence $t^*$ to the source sentence $s$ compared to the top-$k$ extracted-and-edited sentences in the target language space. Intuitively, we view the top-$k$ extracted-and-edited sentences as the anchor points to locate a probable region in the target language space, and iteratively improve the source-to-target mapping via the comparative learning scheme.  

Combined with the language modeling constraints as described in Equation~\ref{equ:lm}, the final loss function for training the the translation model $V_{s\to t}$ is defined as:
\begin{equation}
\begin{split}
\label{equ:generator}
\mathcal{L}_{s\to t}(\theta_{enc},\theta_{dec}|\theta_R) = & \omega_{lm}\mathcal{L}_{lm}(\theta_{enc},\theta_{dec}) + \\ & \omega_{com}\mathcal{L}_{com}(\theta_{enc}|\theta_R),
\end{split}
\end{equation}
where $\omega_{lm}$ and $\omega_{com}$ are hyper-parameters weighing the importance of the language modeling and the comparative learning.

\paragraph{Adversarial Objective}
Meanwhile, we need to learn a good evaluation network $R$ to transform sentence embedding of the shared encoder into the comparable space. The evaluation network $R$ is also shared by two languages to ensure a strong connection between two language spaces.
Inspired by adversarial learning~\cite{goodfellow2014generative}, we can view our translation system as a ``generator" that learns to generate a good translation with a higher similarity score than the extracted-and-edited sentences, and the evaluation network $R$ as a ``discriminator" that learns to rank the extracted-and-edited sentences (real sentences in the target language space) higher than the translated sentences. Thus, we have the following objective function for the evaluation network $R$:
\begin{equation}
\label{equ:discrimintor}
\mathcal{L}_{R}(\theta_{R}) = -\mathbb{E}_{t'\in M'}(\log P(t^{'}|s,M')).
\end{equation}
Based on \autoref{equ:generator} and \ref{equ:discrimintor}, the final adversarial objective is defined as 
\begin{equation}
\begin{split}
 &\min_{\theta_{enc}, \theta_{dec}}\max_{\theta_R}\mathcal{L}(\theta_{enc}, \theta_{dec},\theta_R) \\
 &= -\mathcal{L}_{R}(\theta_{R}) + \mathcal{L}_{s\to t}(\theta_{enc},\theta_{dec}|\theta_R) ,
\end{split}
\end{equation}
where the translation model $V_{s\to t}$ and the evaluation network $R$ play the two-player mini-max game. We evenly alternately update between the encoder-decoder translation model and the evaluation network. The detailed training procedure is described in Algorithm~\ref{algo:training}.

\begin{algorithm}[t]
\small
\caption{The algorithm of our unsupervised NMT system with \emph{extract-edit} approach.}\label{algo:training}
Given two monolingual corpora, source $\mathcal{S}$ and target $\mathcal{T}$;\\
\textbf{Initialization} as in Section~\ref{subsec:init};\\
\textbf{Language Modeling} as in Section~\ref{subsec:lm} to obtain the initialized translation model $V^{(0)}_{s\to t} = enc^{(0)}\circ dec^{(0)}$;\\
 \For{n $\leftarrow$ 1 \KwTo N}{
    Given a source sentence $s$;\\
    \textbf{Extract} the top-$k$ target sentences as the set $M$;\\
    \textbf{Edit} the sentences in $M$ to obtain the set $M'$;\\
    \textbf{Update} the evaluation network $R: \theta_R \leftarrow \text{argmin} \mathcal{L}_R$;\\
    \textbf{Update} the shared encoder and decoder $R$: $\theta_{enc}, \theta_{dec} \leftarrow \text{argmin}\mathcal{L}_{s\to t}$;\\
    \textbf{Update} the translation model: $V^{(n+1)}_{s\to t} = enc^{(n)}\circ dec^{(n)}$;\\
 }
\textbf{return} $V^{(N+1)}_{s\to t} = enc^{(N)}\circ dec^{(N)}$.
\end{algorithm}

\subsection{Model Selection}
\label{subsubsec:selection}
In the fully unsupervised setting, we do not have access to parallel sentence pairs. Thus, we need to find a criterion correlated with the translation quality to select hyper-parameters.
For a neural translation model $V_{s\to t}$, we propose the following criterion $D_{s\to t}$ to tune the hyper-parameters:
\begin{equation}
D_{s\to t} = \mathbb{E}_{s\in\mathcal{S}}[\mathbb{E}(\log P(t^{*} | s,M'))],
\end{equation}
where $t^{*}=V_{s\to t}(s)$. Basically, we choose the hyper-parameters with the maximum expectation of the ranking scores of all translated sentences.

\section{Experiments}
\label{sec:exp}

\begin{table*}[ht]
\small
\begin{center}
\begin{tabular}{l|c|c|c|c}
\toprule 
\textbf{Model} & \textbf{\emph{en}$\to$\emph{fr}} & \textbf{\emph{fr}$\to$\emph{en}} & \textbf{\emph{en}$\to$\emph{de}} & \textbf{\emph{de}$\to$\emph{en}} \\
\midrule
\multicolumn{5}{@{}c@{}}{LSTM Cell} \\
\midrule
\newcite{lample2018phrase} & 24.28 \en& 23.74 \en& 14.71 \en& 19.60 \en\\
Ours (Top-$1$ Extract) & 24.43 \up{+0.15}& 23.90 \up{+0.16}& 14.54 \dn{-0.17}& 19.49 \dn{-0.11}\\
Ours (Top-$1$ Extract + Edit) & 24.54 \up{+0.26}& 24.08 \up{+0.34}& 14.63 \dn{-0.08}& 19.57 \dn{-0.03}\\
Ours (Top-$10$ Extract) & 26.12 \up{+1.84}& 25.83 \up{+2.09}& 17.01 \up{+2.30}& 21.40 \up{+1.80}\\
Ours (Top-$10$ Extract + Edit) & \bf{26.97} \up{+2.69}& \bf{26.66} \up{+2.92}& \bf{17.48} \up{+2.77}& \bf{21.93} \up{+2.33}\\
\midrule
\multicolumn{5}{@{}c@{}}{Transformer Cell} \\
\midrule
\newcite{lample2018phrase} & 25.14 \en& 24.18 \en& 17.16 \en& 21.00\en\\
Ours (Top-$1$ Extract) & 25.30 \up{+0.16}& 24.23 \up{+0.05}& 17.12 \dn{-0.04}& 21.06 \up{+0.06}\\
Ours (Top-$1$ Extract + Edit) & 25.44 \up{+0.30}& 24.36 \up{+0.18}& 17.14 \dn{-0.02}& 21.10 \up{+0.10}\\
Ours (Top-$10$ Extract) & 26.91 \up{+1.77}& 25.64 \up{+1.46}& 19.11 \up{+1.95}& 22.84 \up{+1.84}\\
Ours (Top-$10$ Extract + Edit) & \bf{27.56} \up{+2.42}& \bf{26.90} \up{+2.72}& \bf{19.55} \up{+2.39}& \bf{23.29} \up{+2.29}\\
\bottomrule
\end{tabular}
\end{center}
\vspace{-4ex}
\end{table*}

\begin{table*}[ht]
\small
\caption{\label{tab:translation}The experimental results on all four language pairs and directions. The results are evaluated with BLEU metric on \emph{newstest} 2014 for \emph{en}$\leftrightarrow$\emph{fr}and \emph{newstest} 2016 for \emph{en}$\leftrightarrow$\emph{de}, \emph{en}$\leftrightarrow$\emph{ro} and \emph{en}$\leftrightarrow$\emph{ru}. The \up{+} and \dn{-} stand for performance gains and loss separately compared with baseline models with the same NMT cells.}
\begin{center}
\begin{tabular}{l|c|c|c|c}
\toprule 
\textbf{Model} & \textbf{\emph{en}$\to$\emph{ro}} & \textbf{\emph{ro}$\to$\emph{en}} & \textbf{\emph{en}$\to$\emph{ru}} & \textbf{\emph{ru}$\to$\emph{en}} \\
\midrule
\multicolumn{5}{@{}c@{}}{LSTM Cell} \\
\midrule
\newcite{lample2018phrase} & 19.65 \en& 18.52 \en& 6.24 \en& 7.83 \en\\
Ours (Top-$1$ Extract) & 19.73 \up{+0.08}& 18.56 \up{+0.04}& 6.32 \up{+0.08}& 7.99 \up{+0.16}\\
Ours (Top-$1$ Extract + Edit) & 19.81 \up{+0.16}& 18.69 \up{+0.17}& 6.44 \up{+0.20}& 8.12 \up{+0.29}\\
Ours (Top-$10$ Extract) & 21.57 \up{+1.92}& 20.32 \up{+1.80}& 8.87 \up{+2.63}& 9.76 \up{+1.93}\\
Ours (Top-$10$ Extract + Edit) & \bf{22.08} \up{+2.43}& \bf{20.83} \up{+2.31}& \bf{9.35} \up{+3.11}&\bf{10.21} \up{+2.38}\\
\midrule
\multicolumn{5}{@{}c@{}}{Transformer Cell} \\
\midrule
\newcite{lample2018phrase} & 21.18 \en& 19.44 \en& 7.98 \en& 9.09 \en\\
Ours (Top-$1$ Extract) & 21.15 \dn{-0.03}& 19.52 \up{+0.08}& 8.03 \up{+0.05}& 9.20 \up{+0.11}\\
Ours (Top-$1$ Extract + Edit) & 21.23 \up{+0.05}& 19.59 \up{+0.15}& 8.16 \up{+0.18}& 9.28 \up{+0.19}\\
Ours (Top-$10$ Extract) & 23.04 \up{+1.86}& 21.43 \up{+1.99}& 10.24 \up{+2.26}& 12.29 \up{+3.20}\\
Ours (Top-$10$ Extract + Edit) & \bf{23.31} \up{+2.13}& \bf{21.60} \up{+2.16}& \bf{11.07} \up{+3.09}& \bf{12.72} \up{+3.63}\\
\bottomrule
\end{tabular}
\end{center}
\vspace{-1ex}
\end{table*}

\subsection{Datasets}
\label{subsec:data}
We consider four language pairs: English-French (\emph{en-fr}), English-German (\emph{en-de}), English-Russian (\emph{en-ru}) and English-Romanian (\emph{en-ro}) for evaluation.
We use the same corpora as in \citet{lample2018phrase} for these languages for fair comparison. For English, French, German and Russian, all the available sentences are used from the WMT monolingual News Crawl datasets from years 2007 through 2017. As for Romanian, we combine the News Crawl dataset and WMT'16 monolingual dataset. The translation results are evaluated on newstest 2014 for \emph{en-fr}, and newstest 2016 for \emph{en-de}, \emph{en-ro} and \emph{en-ru}.

\subsection{Implementation Details}
\label{subsec:detail}
We follow previous methods~\cite{koehn2007moses,lample2018phrase} to initialize our models.

\paragraph{Initialization}
\label{subappendix:init}
We use Moses scripts~\cite{koehn2007moses} for tokenization. While the system requires cross-lingual BPE embeddings to initialize the shared lookup table for related languages, we set the number of BPE codes as $60,000$. Following the previous preprocessing protocol~\cite{lample2018phrase}, the embeddings are then generated using fastText~\cite{bojanowski2016enriching} with an embedding dimension of 512, a context window of size 5 and 10 negative samples.

\paragraph{Model Structure}
\label{subappendix:structure}
In this work, the NMT models can be built upon long short-term memory (LSTM)~\cite{hochreiter1997long} and Transformer~\cite{vaswani2017attention} cells. For LSTM cells, both the encoder and decoder have $3$ layers. As for Transformer, we use $4$ layers both in the encoder and the decoder. As for both LSTM and Transformer, all encoder parameters are shared across two languages. Similarly, we share all decoder parameters across two languages. Both two model structure are optimized using Adam~\cite{kingma2014adam} with a batch size of $32$.  The rate for LSTM cell is $0.0003$ while Transformer's is set as $0.0001$. The weights in Equation~\ref{equ:generator} are $\omega_{lm}=\omega_{ext}=1$. The $\lambda$ for calculating ranking scores is $0.5$. As for the evaluation network $R$, we use a multilayer perceptron with two hidden layers of size $512$. For efficient nearest neighbor search in the extracting step, we use the open-source Faiss library~\cite{JDH17}\footnote{\url{https://github.com/facebookresearch/faiss}}. We calculate the similarity of sentences in each episode instead of each batch for computational efficiency. At decoding time, sentences are generated using greedy decoding.

\subsection{Results and Analysis}
\label{subsec:result}
\begin{table*}[ht]
\small
\caption{\label{tab:extraction}The experimental results of parallel sentence mining on the \emph{newstest} 2012 \emph{en} $\to$ \emph{fr} translation dataset with different levels of added sentence noises. Metric: The percentage of Hits@$k$.}
\begin{center}
\begin{tabular}{lcccccccc}
\toprule 
\textbf{Noise} & \textbf{Model} & \textbf{Hits@1} & \textbf{Hits@3} & \textbf{Hits@5} & \textbf{Hits@8} & \textbf{Hits@10} & \textbf{Hits@15} & \textbf{Hits@20} \\
\midrule
\multirow{2}{*}{0\%} & Supervised (Upperbound) & 67.3 & 80.7 & 89.9 & 94.5 & 97.1 & 98.7 & 99.3 \\
& Unsupervised (Ours) & 52.2 & 54.6 & 68.8 & 80.2 & 89.1 & 91.8 & 93.3 \\
\midrule
\multirow{2}{*}{50\%} & Supervised (Upperbound) & 64.8 & 78.0 & 86.8 & 91.3 & 95.6 & 97.4 & 99.0 \\
& Unsupervised (Ours) & 46.9 & 49.7 & 62.1 & 73.4 & 83.2 & 87.6 & 89.2 \\
\midrule
\multirow{2}{*}{90\%} & Supervised (Upperbound) & 63.7 & 76.4 & 84.2 & 89.1 & 93.8 & 96.5 & 98.1 \\
& Unsupervised (Ours) & 41.5 & 46.8 & 58.0 & 69.3 & 77.2 & 83.9 & 87.8 \\
\bottomrule
\end{tabular}
\end{center}
\end{table*}

In this study, we aim to validate the effectiveness of extract-edit versus back-translation for unsupervised neural machine translation (NMT), so we set the unsupervised NMT method in \newcite{lample2018phrase} as the baseline because it currently achieves the state-of-the-art performance on all language pairs.\footnote{Note that for a fair comparison, we are not including the results of the unsupervised phrase-based statistical machine translation system (PBSMT). But theoretically, our extract-edit learning framework can be generalized to other types of machine translation systems such as PBSMT.} 
The overall translation results across four language pairs are shown in Table~\ref{tab:translation}. 
In most of the cases, our proposed extract-edit approach can outperform the baseline models trained with back-translation. Our full models (LSTM/Transformer + Top-$10$ Extract + Edit) achieve more than 2 BLEU points improvement consistently across all the language pairs. Especially, on the \emph{ru} $\to$ \emph{en} translation with the Transformer cell, our full model surpasses the baseline score by $3.63$ BLEU points. These results validate the effectiveness of our approach and indicate that the proposed extract-edit learning framework can learn a better mapping and alignment between language spaces than back-translation.

However, if extracting only top-$1$ target sentence in our approach, the performances are not always improved (e.g., \emph{en} $\to$ \emph{de}, \emph{de} $\to$ \emph{en}, and \emph{en} $\to$ \emph{ro}). Besides, \textit{Top-10 Extract + Edit} models consistently outperforms \textit{Top-1 Extract + Edit}.
This is because more extracted-and-edited sentences lead to a higher recall, so more useful information will be used to guarantee the translation quality. The comparative translation loss can avoid the model suffering from the noise while taking advantage of more information. In other words, it is more likely to project the source sentence into the probable region in the target language space with more sentences serving as the anchor points, and the comparative learning scheme iteratively approximates towards more accurate target points.
This highlights the importance of the extraction number $k$, which we further discuss next.

\subsection{Ablation Study}
\label{subsec:ablation}

\begin{figure}[t]
\centering
\small
\includegraphics[width=0.9\textwidth]{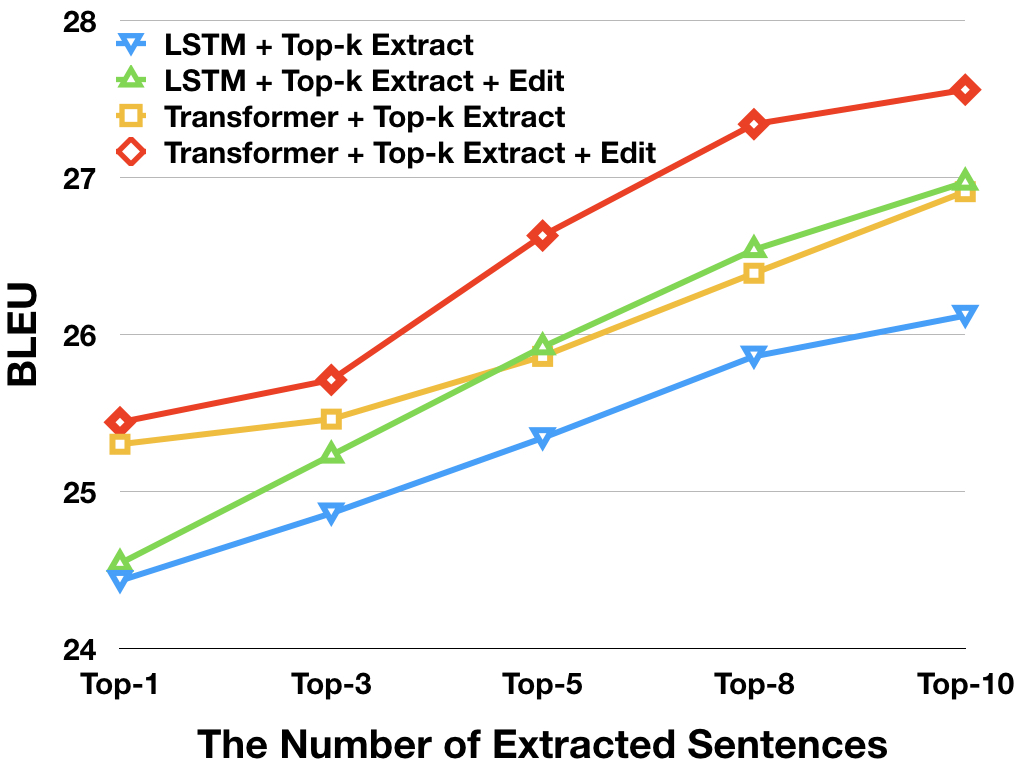}
\caption{The effect of the number $k$ of the extracted sentences in our approach on \emph{en}$\to$\emph{fr} translation.}
\label{fig:topk} 
\end{figure}

\paragraph{The Effect of Extraction Number $k$}
\label{subsubsec:knumber}
As shown in Table~\ref{tab:translation}, the number $k$ of the extracted-and-edited sentences plays a vital role in our approach. Thus for a more intuitive overview of its impact, we further train and evaluate multiple models with $k=1,3,5,8,10$ on \emph{en} $\to$ \emph{fr} translation task. The detailed results are shown in Figure~\ref{fig:topk}.
This study shows that the translation performance of our approach is indeed improving as $k$ increases. 
Because as we analyze above, larger $k$ ensure a higher recall and thus more critical semantic information can be utilized to assist the translation.
Besides, the diversity of extracted-and-edited sentences can potentially provide a more accurate localization of the probable region where the target sentence should be.
Although we can infer that the models would perform even better with $k>10$ from Figure~\ref{fig:topk}, more computational resources will be required for that and we already observe a decelerated growth of BLEU scores from $k=8$ to $k=10$. Therefore, in this paper, we set $k=10$ for the full models. 

\paragraph{The Quality of Extraction Model}
\label{subsubsec:test}
In this section, we quantitatively evaluate the unsupervised extraction part of our model and compare it with the state-of-the-art supervised extraction model. 
Following \citet{gregoire2018extracting}, we train a fully supervised parallel pair extraction model, where two Bi-LSTMs are implemented to encode sentences of two languages, and a feed-forward network is followed to culminate in a sigmoid output layer.
The model is trained with around $500,000$ English-French parallel sentence pairs sampled from Europarl corpus~\cite{koehn2005europarl}.
As for our unsupervised extraction model, we directly use the jointly trained extraction part in our framework to extract the potential parallel sentences based on the scores computed by Equation~\ref{equ:cosine}.
For evaluation, we sample $1,000$ parallel sentences from the \emph{newstest} 2014 corpus and create three test sets with a noise ratio 0\%, 50\%, and 90\% to simulate noisy real-world data. We report Hits@$k$ results, which shows the percentage of the golden parallel sentences appear within the top-$k$ place.

The detailed results are shown in Table~\ref{tab:extraction}. Although our extraction model structure is different from the supervised extraction model, it can still give us a good insight into the upperbound and gap of performance. 
We can observe a noticeable gap between unsupervised and supervised methods, but the gap is narrowing as the rank increases. Meanwhile, in our unsupervised method, the performance grows quickly when $k\leq10$. From \autoref{tab:extraction} we also notice that $k=10$ is a sweet point, where the accuracy is high and the computational cost is relatively acceptable.

\begin{table}
\small
\caption{\label{tab:retrain}The performance of the unsupervised NMT systems with different learning objectives on \emph{en} $\to$ \emph{fr} \emph{newstest} 2014.}
\begin{center}
\begin{tabular}{llc}
\toprule 
Cell & \textbf{Learning} & \textbf{BLEU}\\
\midrule
\multirow{2}*{LSTM} & MLE Loss & 12.40 \\
~ & Comparative Loss & 24.54 \\
\midrule
\multirow{2}*{Transformer} & MLE Loss & 14.15 \\
& Comparative Loss & 25.44 \\
\bottomrule
\end{tabular}
\end{center}
\vspace{-1ex}
\end{table}

\paragraph{The Effect of Comparative Translation}
\label{subsubsec:loss_effect}
Finally, we aim to roughly evaluate the effect of the proposed comparable translation loss in our model. Thus, we compare our model with a two-staged NMT system, where we extract and edit the parallel pairs and retrain the NMT system with the standard maximum likelihood estimation (MLE) loss in a supervised way (by taking the extracted-and-edited sentences as the ground-truth targets).
We compare the performance on the \emph{en} $\to$ \emph{fr} \emph{newstest} 2014 dataset, and the results are shown in Table~\ref{tab:retrain}. We can observe that with the MLE loss, the translation performance will drop nearly $50\%$. The results indirectly reflect that the NMT systems are sensitive to noises in the training datasets. Meanwhile, it demonstrates by treating extracted-edit sentences as pivotal points instead of ground truth, our proposed comparative translation loss can avoid the NMT model suffers from the noise. 

\subsection{Discussion}
\label{subsec:discussion}
Although our extract-edit approach can achieve better performance than the back-translation mechanism, it is still worth mentioning that our approach has more strict constraints on the domains of the source and target corpus. The extract-edit approach will work well when there is information overlap in the two language spaces. When there is little overlap in terms of domains, it will be much harder to find a good cluster of initial candidates, which may also complicate the editing process. As for the back-translation mechanism, it requires less overlap in terms of the language spaces because the language priors can be learnt in any domains. However, the corpus with matching domains can be easily obtained nowadays (e.g., Wikipedia and the news articles), which makes our extract-edit approach still widely applicable.

\section{Related Work}
\label{sec:related}
\paragraph{Unsupervised NMT}
The current NMT systems~\citep{sutskever2014sequence,cho2014properties,bahdanau2014neural,gehring2017convolutional,vaswani2017attention} are known to easily over-fit and result in an inferior performance when the training data is limited \cite{koehn2017six,isabelle2017challenge,sennrich2016grammatical}. Many research efforts have been spent on how to utilize the monolingual data to improve the NMT system when only limited supervision is available~\cite{gulcehre2015using,sennrich2015improving,he2016dual,zhang2016exploiting,yang2018unsupervised}. 
Recently, \newcite{lample2017unsupervised,artetxe2017unsupervised,lample2018phrase} make encouraging progress on unsupervised NMT structure mainly based on initialization, denoising language modeling, and back-translation. However, all these unsupervised models are based on the back-translation learning framework to generate pseudo language pairs for training. Our work leverages the information from real target language sentences.

\paragraph{Comparable Corpora Mining}
Comparable corpora mining aims at extracting parallel sentences from comparable monolingual corpora such as news stories written on the same topic in different languages. Most of the previous methods align the documents based on metadata and then extract parallel sentences using human-defined features~\cite{W02-1037,P06-1011,hewavitharana2011extracting}.
Recent neural-based methods~\cite{chu2016parallel,grover2017bilingual,gregoire2018extracting} learn to identify parallel sentences in the semantic spaces. 
However, these methods require large amounts of parallel sentence pairs to train the systems first and then test the performance on raw comparable corpora, which does not apply to languages with limited resources. Instead, we explore the corpora mining in an unsupervised fashion and propose a joint training framework with machine translation.

\paragraph{Retrieval-Augmented Text Generation}
Our work is also related to the recent work on applying retrieval mechanisms to augment text generation, such as image captioning~\cite{kuznetsova2013generalizing,mason2014domain}, dialogue generation~\cite{song2016two,yan2016learning,wu2018response} and style transfer~\cite{lin2017adversarial,li2018delete}. Some editing-based models~\cite{guu2018generating,hashimotoretrieve} are proposed to further enhance the retrieved text. Recent work in machine translation~\cite{gu2017search} augments an NMT model with sentence pairs retrieved by an off-the-shelf search engine. However, these methods are two-staged with supervised retrieval first. In our work, the extracted-edited sentences are not directly used as the ground truth to train the translation model. Instead, we view these sentences as pivotal points in the target language space and further we propose a comparative translation loss to train the system in a fully unsupervised way.

\section{Conclusion}
\label{sec:con}
In this paper, we propose an \emph{extract-edit} approach, an effective alternative to the widely-used back-translation in unsupervised NMT. Instead of generating pseudo language pairs to train the systems with the reconstruction loss, we design a comparative translation loss that leverages real sentences in the target language space. Empirically, our method advances the previous state-of-the-art NMT systems across four language pairs using the monolingual corpora only. Theoretically, we believe the extract-edit learning framework can be generalized to other types of unsupervised machine translation systems and even some other unsupervised learning tasks.

\section*{Acknowledgments}
The authors would like to thank the anonymous reviewers for their thoughtful comments. The work was supported by the Facebook Low Resource Neural Machine Translation Research Award. The authors are solely responsible for the contents of the paper, and the opinions expressed in this publication do not reflect those of the funding agencies.

\bibliography{naaclhlt2019}
\bibliographystyle{acl_natbib}

\end{document}